# A Dataset for Addressing Patient's Information Needs related to Clinical Course of Hospitalization

Sarvesh Soni, Dina Demner-Fushman
Division of Intramural Research
National Library of Medicine, National Institutes of Health, Bethesda, MD, USA
sarvesh.soni@nih.gov

# Abstract

**Background:** Patients have distinct information needs related to their hospitalization that can be fulfilled using clinical evidence from electronic health records (EHRs). Advanced artificial intelligence (AI) systems have the potential to fulfill these needs; however, robust, high-quality datasets are essential to evaluate the factual accuracy and relevance of AI-generated responses. To our knowledge, no existing dataset captures patient information needs in the context of their EHRs.

**Methods:** We developed a novel, expert-annotated dataset, ArchEHR-QA, based on real-world patient cases from intensive care unit and emergency department settings. The cases comprise questions posed by patients to public health forums, clinician-interpreted counterparts, relevant clinical note excerpts with sentence-level relevance annotations, and clinician-authored answers. To establish benchmarks for grounded EHR question answering (QA), we evaluated three open-weight large language models (LLMs)—Llama 4, Llama 3, and Mixtral—using three prompting strategies: (1) generating answers with citations to clinical note sentences, (2) generating answers before citations, and (3) generating answers from filtered citations. We assessed performance based on two metrics: *Factuality* (overlap between cited note sentences and ground truth relevance annotations) and *Relevance* (textual and semantic similarity between system and reference answers).

**Results:** The final dataset contains 134 patient cases. Across all prompting strategies and models, the answer-first prompting approach consistently yielded the highest performance. Llama 4 outperformed the other models across strategies, achieving the best overall score using the answer-first method. Manual error analysis supported the automated findings and highlighted common issues, including omission of critical clinical evidence and incorporation of contradictory or hallucinated information.

**Conclusions:** We introduce a first-of-its-kind dataset to support the development and evaluation of patient-centered EHR QA systems. Our benchmarking results establish performance baselines and emphasize the need for further advancements to improve the factual consistency and relevance of AI-generated responses in clinical contexts.

# Introduction

Question answering (QA) is an organic way to interact with complex information systems such as electronic health records (EHRs)[1], where a QA system responds to user questions with exact answers. The major focus of existing EHR QA work has been on addressing clinician information needs[2], with datasets for system development and evaluation largely prioritizing these requirements[3]. However, with the increasing patient-involvement in their care[4–8], there is a need for targeted EHR QA research that incorporates the unique needs of patients from their health records[9,10]. To this end, datasets play an important role in developing and evaluating tailored artificial intelligence (AI) systems and, thus, the datasets must be representative of the target end-user's needs[11], i.e., patients.

Despite growing interest in patient-centered care, most work on addressing the health information needs of patients uses general health resources[12]. This disconnect often results in systems that fail to incorporate the context that prompts the patients' questions, i.e., EHRs, contributing to non-patient-centered solutions[13]. EHRs document important clinical evidence about the care provided to patients. However, the rationale for drafting these documents often differs from the primary reasons patients access them[14]. This necessitates building assistive technologies to contextualize EHR information to address patients' information needs.

Moreover, the volume of patient requests for medical information through patient portals is rising, contributing to desktop medicine and increasing clinician burden[15]. Most existing studies on automated responses to patient messages do not incorporate critical contextual information from EHRs[16–20]. Among studies that use EHR content, none provide comprehensive evaluations of how effectively the generated responses leverage this clinical context[21,22].

Grounding is crucial in AI applications in medicine, as it ensures that AI models are anchored to accurate, contextually relevant, real-world clinical data. This is particularly important when the intended audience lacks clinical expertise[23]. To effectively design and evaluate grounded QA systems, a representative dataset and evaluation framework is essential[24].

In this work, we propose a benchmark dataset, ArchEHR-QA, to evaluate grounding capabilities of models for responding to patient-initiated queries. The proposed dataset consists of patient-initiated questions posted in public domain, the corresponding clinician-interpreted questions, their answers annotated in the EHRs, and clinician-generated free-text answers to the questions. We collect true patient health information needs by aligning real-world health forum messages to publicly accessible real EHRs. To our knowledge, this is the first public dataset that encapsulates patient questions and relevant clinical evidence from EHRs. We further provide an evaluation framework to assess two critical aspects of a grounded EHR QA system: does it identify relevant information from given clinical evidence and does it use this information in responding to user queries.

**Table 1.** Example annotated case from the dataset with a sample system-generated response.

| |
|---|
| **Patient Question** *(underlined are the areas of focus)* |
| Took my 59 yo father to ER ultrasound discovered he had an aortic aneurysm. <u>He had a salvage repair (tube graft).</u> Long surgery / recovery for couple hours then removed packs. <u>why did they do this surgery?????</u> After this time he spent 1 month in hospital now sent home. |
| **Clinician Question** *(interpreted from the patient question)* |
| Why did they perform the emergency salvage repair on him? |
| **Clinical Note Excerpt** *(sentences numbered for grounding)* – 1 and 2 are essential; 3 is supplementary; all others are not-relevant |
| *1:* He was transferred to the hospital on 2025-1-20 for emergent repair of his ruptured thoracoabdominal aortic aneurysm. *2:* He was immediately taken to the operating room where he underwent an emergent salvage repair of ruptured thoracoabdominal aortic aneurysm with a 34-mm Dacron tube graft using deep hypothermic circulatory arrest. *3:* Please see operative note for details which included cardiac arrest x2. *4:* Postoperatively he was taken to the intensive care unit for monitoring with an open chest. *5:* He remained intubated and sedated on pressors and inotropes. *6:* On 2025-1-22, he returned to the operating room where he underwent exploration and chest closure. *7:* On 1-25 he returned to the OR for abd closure JP/ drain placement/ feeding jejunostomy placed at that time for nutritional support. *8:* Thoracoabdominal wound healing well with exception of very small open area mid wound that is @1cm around and 1/2cm deep, no surrounding erythema. *9:* Packed with dry gauze and covered w/DSD. |
| **Clinician Answer** *(with citations to note sentences) – some sentences may be unsupported* |
| The patient needed emergency salvage repair for his aortic aneurysm because the aorta had ruptured [1]. This rupture is something that needs to be repaired immediately or the patient will die. The patient needed a tube graft for the repair [2]. An aortic aneurysm is a very serious diagnosis and repair surgery is not something that can be put on hold. Additionally, it appears this surgery was absolutely necessary because the patient arrested twice during the operation [3]. |
| **System-generated Answer** *(with citations to note sentences) – some sentences may be unsupported* |
| They performed the emergency salvage repair because the patient had a ruptured thoracoabdominal aortic aneurysm. |1,2| The surgery was necessary to address the rupture, which required immediate attention, and involved a 34-mm Dacron tube graft using deep hypothermic circulatory arrest. |2| The procedure was done to save the patient's life after he experienced cardiac arrest twice during the operation. |3| |

# Methods

## Data

There is no public resource containing both patient-initiated questions and their EHRs. Thus, we created ArchEHR-QA by aligning two different publicly available resources for patient questions and EHRs (Figure 1). To incorporate true patient information needs, we derive questions using public health forums (specifically, HealthCareMagic, now Ask A Doctor-24x7)[25], one of the most popular internet venues for patients to seek information where patients pose their information needs as discussion posts[26,27]. To associate relevant medical context to the questions, we find clinical evidence using discharge summaries from publicly accessible EHR data (specifically, MIMIC-III and MIMIC-IV databases which contain deidentified patient records[28,29]).

## Alignment

Health forum posts were filtered to reflect scenarios in which patients had recently been discharged from the hospital—a common context for post-encounter online information seeking[30]. The posts containing keywords such as "admission", "emergency room", and "ICU" (see Supplementary Materials for a complete list) were selected and reviewed by SS (clinical informaticist) and DDF (MD and computational linguist). For each reviewed forum post, a set of candidate relevant discharge summaries (notes) were retrieved from the EHR. Due to the differences in vocabulary between patient-authored posts and clinician- authored notes (e.g., *"pelvic"* vs *"inferior perineal"*), we augmented the text in both sources with standardized terminology. For forum posts, we appended ICD code descriptions automatically generated by prompting the Llama 3.3 70B[31] large language model (LLM; prompt details available in the supplementary materials). For clinical notes, we appended the corresponding ICD code descriptions associated with the patient's admission (available as part of the structured data in the MIMIC databases). The expanded posts were used to retrieve candidate notes using BM25 algorithm[32]. The top 10 notes for each post were manually reviewed by SS and DDF to select only the ones that may be used as evidence to satisfy the information needs expressed in the forum post. For each selected note, the corresponding forum post was minimally modified to match the surrounding details (e.g., changing *"mother"* to *"father"*) to ensure its alignment with the note. We never modified the underlying information needs in the post or the associated note content. Annotation guidelines are provided in Supplementary Material.

## Question and Answer Creation

For each aligned forum post-note pair, SS, DDF, and a licensed clinician annotated the query-related focus areas in the post and created a clinician's version of the patient's query—this is a question that a clinician would interpret and respond to (Table 1). Further, each sentence in the clinical note was manually annotated with a *"relevance"* label, indicating its importance in answering the question: *"essential"* (must be used), *"supplementary"* (provides support but not necessary), or *"not-relevant"*. Due to the considerable length of the notes, we manually curated *note excerpts* by excluding paragraphs composed entirely of sentences deemed *"not-relevant"*, while retaining those that provided important contextual information (e.g., section headers such as *"Brief Hospital Course"*). Lastly, we tasked licensed clinicians (hired through Centaur Labs) to review and update the relevance labels for the note sentences and compose 75-word (approximately 5-sentence) *ground truth answers* to the questions.

## Validation

All annotations were reviewed and revised as necessary by a second annotator. Alignment and sentence labeling was performed by the authors and reviewed by a licensed clinician, who revised the annotations as needed and authored an answer. Lastly, the clinician-written answer was reviewed by DDF, revising as needed. We report the annotation agreements for all stages.

We performed baseline experiments for the task of automatically generating an answer to the input question (using both patient and clinician versions) with citations to the specific sentences in the input note excerpt. Based on existing literature[33,34], we restricted the output answer length to 75 words (or approximately 5 sentences). The baseline approaches (Figure 2) include a heuristic method that selects the most semantically similar note sentences using cosine similarity over MiniLM semantic embeddings[35] and restricts the response to the top-ranked sentences within the word limit.

In addition to the heuristic baseline, we evaluated three prompting strategies using LLMs, each reflecting a distinct approach to citation integration: (1) generating answers with citations embedded directly in the output, (2) generating answers first followed by the addition of citations in a subsequent step, and (3) generating answers using only a specified set of citations as input. These strategies were designed to evaluate how citation context and ordering influence the model's ability to produce factual and relevant responses.

All prompts are structured into four components: (a) an *input section* containing the patient and clinician questions; (b) *task-specific instructions*, which defined the objective of the prompt (e.g., for the *"Answer & Cite"* task, instructions explicitly directed the model to generate a response with inline citations); (c) *formatting instructions*, which specified the required output structure (e.g., for the *"Answer & Cite"* task, responses were to be formatted in markdown, with each answer sentence presented on a separate line followed by the corresponding citations); and (d) a *feedback section* providing error-specific guidance based on prior output. During prompt development, we observed several correctable errors, such as failing to include citations or format content on separate lines. To address these issues, we implemented an iterative prompting strategy, retrying each prompt up to five times. At each iteration, the model received programmatically generated feedback tailored to the specific formatting or content errors identified in the preceding response.

We benchmarked the zero-shot performance of three popular LLMs on ArchEHR-QA: Llama 3.3 70B[31], Llama 4 17B 16E[36], and Mixtral 8x22B[37]. These models represent top-tier performance among open-weight LLMs that can be run locally. The Llama models were developed by Meta AI, while the Mixtral model was released by Mistral AI. Llama 3.3 employs a standard transformer architecture, whereas Llama 4 and Mixtral leverage a mixture-of-experts (MoE) design to improve computational efficiency and performance.

## Evaluation

System-generated responses are evaluated along two dimensions: *Factuality* (use of clinical evidence for grounding) and Relevance (similarity to the ground-truth answer).

*Factuality* is measured using an F1 Score between the sentences cited as evidence in the system-generated answer (which are treated as predicted essential note sentences) and the ground truth relevance labels for note sentences. We define two versions of *Factuality:* *"Essential-only"* and *"Essential + Supplementary"*. In the *"Essential-only"* definition, only sentences labeled as essential in the ground truth count as positives. In the *"Essential + Supplementary"* version, ground truth sentences labeled as either essential or supplementary are counted as positives (penalizing the system for failing to cite either, but not for including supplementary ones). We report *Factuality* both at the *macro* level (averaging per-case F1 scores) and the *micro* level (aggregating true positives, false positives, and false negatives across all cases). We designate the essential-only micro F1 Score as *Overall Factuality* and use it to calculate the *Overall Score*, as it captures aggregate performance across all instances focusing on the most important note sentences.

*Relevance* is evaluated by comparing the generated answer text to a clinician-authored reference answer using text- and semantics-based metrics: BLEU[38], ROUGE[39], SARI[40], BERTScore[41], AlignScore[42], and MEDCON[43]. Each metric is normalized, and *Overall Relevance* is computed as the mean of these normalized scores. We compute another version of relevance scores by treating the set of essential ground truth sentences (together with the original question) as the reference. This alternative provides a feasible and scalable approximation for evaluating answer quality in the absence of human-written answers.

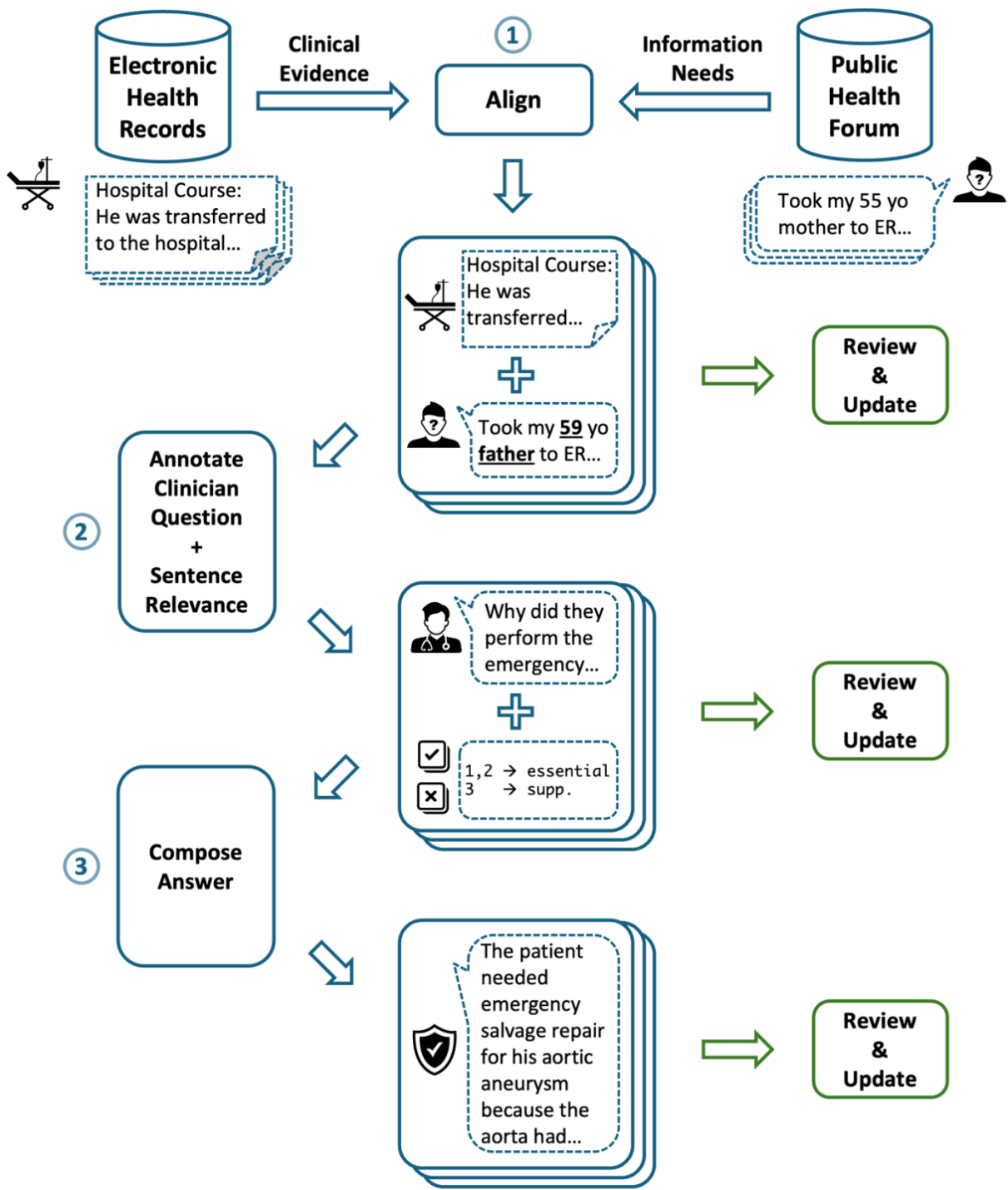

**Figure 1:** Dataset creation workflow. (1) Patient information needs identified from a public health discussion forum are aligned with clinical evidence from an electronic health record database to produce pairs of patient-posed questions and associated clinical notes. (2) Clinician-interpreted questions are created from patient questions and sentence-level relevance labels are annotated in the note. (3) Clinicians compose natural language answers considering the notes. Annotations are reviewed and updated at each stage to ensure high quality.

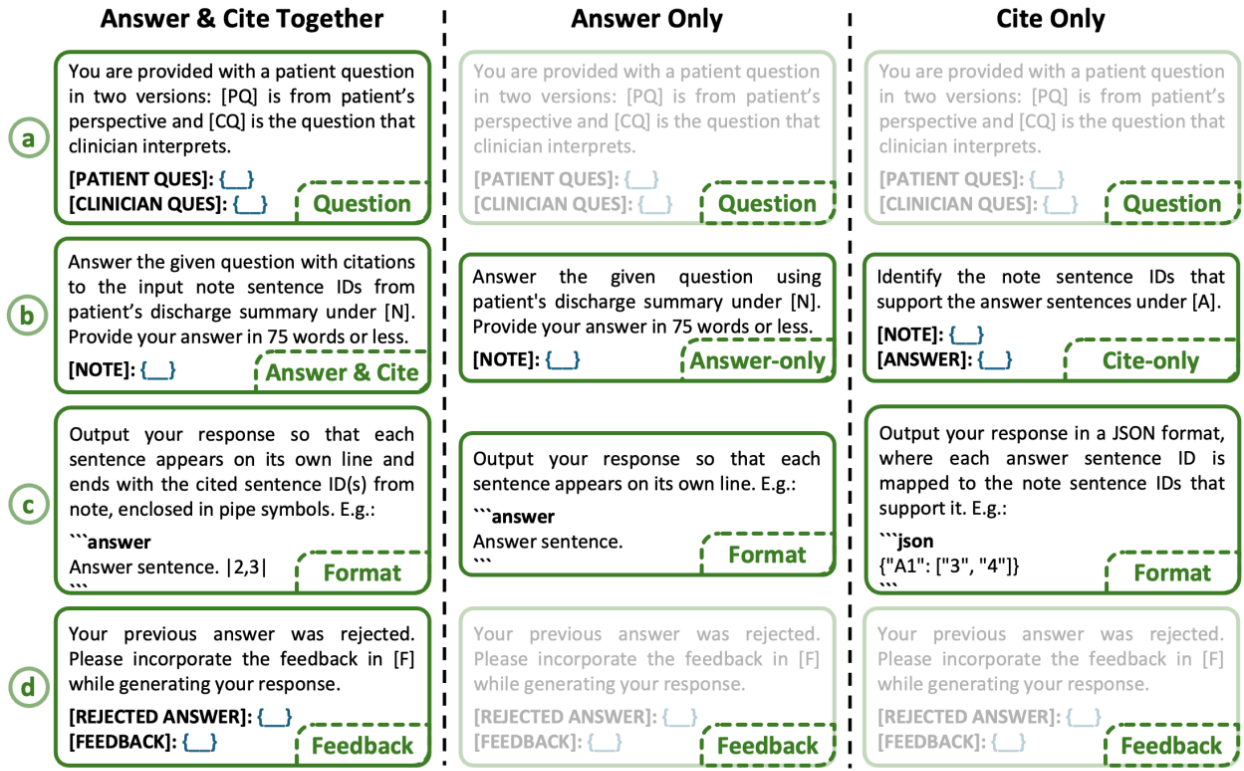

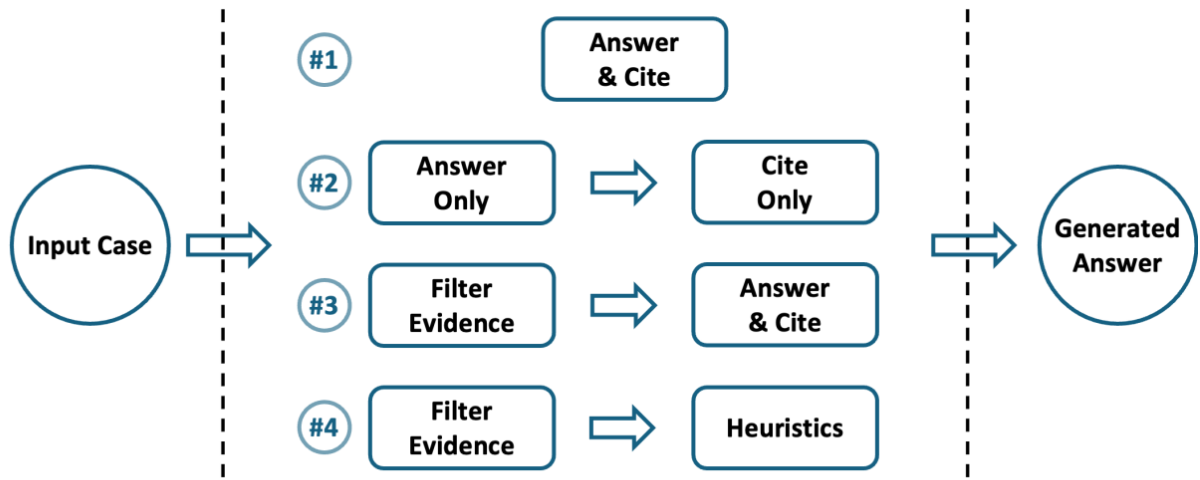

**Figure 2:** Overview of baseline approaches for the task of generating answers with citations. *(A) Prompt structures:* Each prompt is organized into four sections: (a) Input questions; (b) Instructions for the specific task component—*"Answer & Cite"*, *"Answer-only"*, or *"Cite-only"*; (c) Desired response format; and (d) Feedback from prior iterations, if available. Each prompt is executed up to five times to handle errors, after which the final response is output. *(B) Baseline approaches:* Four prompting approaches are evaluated: (1) *"Together"*—generating the answer and citations together; (2) *"Answer First"*—generating the answer first, then citations; (3) *"Evidence First"*—providing the model with filtered evidence sentences (based on cosine similarity to the questions); and (4) *"Heuristics"*—using evidence filtering followed by heuristic selection of supporting sentences until a word limit is reached.

# Results

The proposed dataset contains a total of 134 patient cases—104 using intensive care unit (ICU) notes (MIMIC-III) and 30 using emergency department (ED) notes (MIMIC-IV). Table 2 details patient demographics and case statistics. The average length of patient narratives is 91.3 words (with a standard deviation [SD] of 56.9) while the clinician questions are 10.5 words (SD 3.6) long. The clinician-written answers have an average of 72.6 (SD 3.4) words. The note excerpts contain a mean of 25.7 sentences, 25.7% (6.6) of which are essential and 20.1% (5.2) are supplementary. The mean length of ICU note excerpts exceeds that of ED note excerpts, likely reflecting greater clinical complexity of ICU cases. This is also suggested by the prevalence of ICD codes in each setting. ICU diagnoses predominantly involve heart, kidney, and respiratory failure, which often require intensive and multidisciplinary management, whereas ED diagnoses more often include single-system, episodic, or behavioral conditions such as asthma, apnea, and nicotine dependence.

In the first round of sentence relevance annotations review, clinicians agreed with the existing labels 77.1% (2653/3450) of the time, with the most common edits being changes from essential to supplementary (39.9% [318/797]) or to not-relevant (26.6% [212/797]). In the second round, the reviewer's (DDF) agreement with the clinician-modified labels was 84.7% (2874/3450), with the predominant edit again being from essential to supplementary (56.1% [323/576]). Furthermore, the reviewer (DDF) retained 81.3% of the clinician-written answers, calculated as the proportion of word lemmas retained from the original answers while ignoring punctuations and stop words.

Table 3 summarizes the benchmarking results on a held-out test set of 100 cases (80 from ICU and 20 from ED). The remaining 34 cases were used exclusively for the development of prompts. All the model-based approaches performed the best for *Answer First* variation, followed by *Together* and then *Evidence First* across most evaluation metrics, including the *Overall Factuality* and *Relevance*. Llama 4 consistently achieved superior *Factuality F1* scores across approaches with the best *Overall Factuality* (essential-only micro F1) of 51.8%, followed by Mixtral and Llama 3. In contrast, for *Relevance*, Mixtral outperformed the other models in most instances, achieving the best *Overall Relevance* of 33.8% (compared against human-authored answers), followed by Llama 3 and Llama 4.

On *Factuality*, when the evaluation considered the supplementary sentences as important, F1 scores dropped (driven primarily by a drop in Recall despite gains in Precision). This indicates that models were more effective at identifying the most important information from the clinical notes but struggled to consistently capture supporting details. Notably, macro F1 scores consistently exceeded the micro F1 scores, reflecting consistent performance across cases, despite variations in instance-level scores. Interestingly, *Together* approach often yielded higher precision, though at the expense of a lower recall, leading to modest F1 scores. Also, *Evidence First* consistently produced the poorest F1 scores, even underperforming the Heuristic approach, highlighting that constraining the evidence for the model creates a bottleneck for model performance.

On *Relevance* using human-authored answers, *Evidence First* also underperformed across most metrics. This suggests that pre-selecting evidence negatively affects not only factual correctness but also the relevance to clinician answers. Among models, *Mixtral* attained the highest MEDCON score (40.2%, using *Answer First* approach), indicating strong alignment with clinically relevant medical concepts. Llama 3 attained the highest AlignScore (47.0%, using *Together*), underlining its superior alignment with the clinician answers despite modest performance on other metrics. The *Heuristic* approach showed consistently lower performance across metrics. Notably, its BERTScore dropped to 30.4%, significantly lower than the worst-performing model-based approach (36.9% using *Evidence First* and Llama 4).

When the generated answers were compared to the essential note sentences and the question, the heuristic version achieved the highest scores—an artifact of the relevance metrics' bias toward lexical overlap. Interestingly, the LLM-based variations achieved comparable scores using human answers or notes. There is a statistically significant positive correlation between the overall relevance scores calculated using human answers and notes (Kendall's $\tau$ = 0.362, p < 0.0001, n=1000; comparing 100 individual case scores across 10 model variations), indicating moderate agreement.

**Table 2.** Descriptive statistics of the dataset. ICU: Intensive Care Unit, ED: Emergency Department. Population statistics are in *count (proportion)* format.

| Category | Value | ICU (N=104) | ED (N=30) | All (N=134) |
|---|---|---|---|---|
| **Race or Ethnicity** | White | 74 (71.2%) | 17 (56.7%) | 91 (67.9%) |
| | Black/African American | 13 (12.5%) | 7 (23.3%) | 20 (14.9%) |
| | Hispanic/Latino | 4 (3.8%) | 2 (6.7%) | 6 (4.5%) |
| | Asian | 3 (2.9%) | 2 (6.7%) | 5 (3.7%) |
| | Other | 1 (1.0%) | 1 (3.3%) | 2 (1.5%) |
| | Unknown/Declined | 9 (8.7%) | 1 (3.3%) | 10 (7.5%) |
| **Sex** | Male | 58 (55.8%) | 10 (33.3%) | 68 (50.7%) |
| | Female | 46 (44.2%) | 20 (66.7%) | 66 (49.3%) |
| **Age Group** | 0-30 | 8 (7.7%) | 2 (6.7%) | 10 (7.5%) |
| | 31-40 | 5 (4.8%) | 7 (23.3%) | 12 (9.0%) |
| | 41-50 | 13 (12.5%) | 3 (10.0%) | 16 (11.9%) |
| | 51-60 | 20 (19.2%) | 6 (20.0%) | 26 (19.4%) |
| | 61-70 | 25 (24.0%) | 8 (26.7%) | 33 (24.6%) |
| | 71-80 | 16 (15.4%) | 2 (6.7%) | 18 (13.4%) |
| | 81-89 | 15 (14.4%) | 1 (3.3%) | 16 (11.9%) |
| | 90+ | 2 (1.9%) | 1 (3.3%) | 3 (2.2%) |
| **Top 10 ICD Codes - ICU** | 4019 - Unspecified essential hypertension | 39 (37.5%) | 3 (10.0%) | 42 (31.3%) |
| | 4280 - Congestive heart failure, unspecified | 34 (32.7%) | 0 (0.0%) | 34 (25.4%) |
| | 42731 - Atrial fibrillation | 27 (26.0%) | 0 (0.0%) | 27 (20.1%) |
| | 5849 - Acute kidney failure, unspecified | 26 (25.0%) | 3 (10.0%) | 29 (21.6%) |
| | 51881 - Acute respiratory failure | 26 (25.0%) | 0 (0.0%) | 26 (19.4%) |
| | 5990 - Urinary tract infection, site not specified | 23 (22.1%) | 0 (0.0%) | 23 (17.2%) |
| | 2724 - Other and unspecified hyperlipidemia | 22 (21.2%) | 2 (6.7%) | 24 (17.9%) |
| | 41401 - Coronary atherosclerosis of native coronary artery | 22 (21.2%) | 2 (6.7%) | 24 (17.9%) |
| | 486 - Pneumonia, organism unspecified | 19 (18.3%) | 0 (0.0%) | 19 (14.2%) |
| | 2761 - Hyposmolality and/or hyponatremia | 18 (17.3%) | 1 (3.3%) | 19 (14.2%) |
| **Top 10 ICD Codes - ED** | K219 - Gastro-esophageal reflux disease without esophagitis | 0 (0.0%) | 8 (26.7%) | 8 (6.0%) |
| | J45909 - Unspecified asthma, uncomplicated | 0 (0.0%) | 7 (23.3%) | 7 (5.2%) |
| | G4733 - Obstructive sleep apnea (adult) (pediatric) | 0 (0.0%) | 6 (20.0%) | 6 (4.5%) |
| | Z87891 - Personal history of nicotine dependence | 0 (0.0%) | 6 (20.0%) | 6 (4.5%) |
| | I10 - Essential (primary) hypertension | 0 (0.0%) | 6 (20.0%) | 6 (4.5%) |
| | F419 - Anxiety disorder, unspecified | 0 (0.0%) | 6 (20.0%) | 6 (4.5%) |
| | F329 - Major depressive disorder, single episode, unspecified | 0 (0.0%) | 5 (16.7%) | 5 (3.7%) |
| | D649 - Anemia, unspecified | 0 (0.0%) | 4 (13.3%) | 4 (3.0%) |
| | F17210 - Nicotine dependence, cigarettes, uncomplicated | 0 (0.0%) | 4 (13.3%) | 4 (3.0%) |

|  | E785 - Hyperlipidemia, unspecified | 0 (0.0%) | 4 (13.3%) | 4 (3.0%) |
| --- | --- | --- | --- | --- |
| *Case Statistics* | | | | |
| **Patient Narrative Word Count** | Mean | 90.2 | 94.8 | 91.3 |
| | Median | 72.0 | 90.5 | 76.5 |
| | Std Dev | 61.7 | 36.6 | 56.9 |
| | Min | 33.0 | 54.0 | 33.0 |
| | Max | 440.0 | 192.0 | 440.0 |
| **Clinician Question Word Count** | Mean | 10.6 | 10.2 | 10.5 |
| | Median | 10.0 | 9.0 | 10.0 |
| | Std Dev | 3.6 | 3.8 | 3.6 |
| | Min | 3.0 | 4.0 | 3.0 |
| | Max | 21.0 | 21.0 | 21.0 |
| **Answer Word Count** | Mean | 72.6 | 72.3 | 72.6 |
| | Median | 73.0 | 73.5 | 73.0 |
| | Std Dev | 3.4 | 3.4 | 3.4 |
| | Min | 55.0 | 61.0 | 55.0 |
| | Max | 78.0 | 75.0 | 78.0 |
| **Note Excerpt Word Count** | Mean | 410.2 | 280.7 | 381.2 |
| | Median | 383.5 | 223.0 | 351.5 |
| | Std Dev | 200.1 | 196.8 | 205.9 |
| | Min | 107.0 | 76.0 | 76.0 |
| | Max | 1028.0 | 868.0 | 1028.0 |
| **Mean Note Sentences Count** | All | 27.6 | 19.3 | 25.7 |
| | Essential | 7.0 (25.5%) | 5.2 (26.7%) | 6.6 (25.7%) |
| | Supplementary | 5.9 (21.4%) | 2.6 (13.4%) | 5.2 (20.1%) |
| | Not Required | 14.7 (53.1%) | 11.6 (59.8%) | 14.0 (54.3%) |

**Table 3.** Results from benchmarking experiments. *"Overall Relevance"* is the mean of all normalized relevance scores. *"Overall Score"* is the average of *"Overall Relevance"* and *"Overall Factuality"* (Essential-only Micro F1). All scores are reported as percentages. *Macro:* average per-case F1; *Micro:* aggregate true/false positives and negatives across all cases; *P:* Precision; *R:* Recall; *F1:* F Score; *Llama 3:* Llama 3.3 70B; *Llama 4:* Llama 4 17B 16E; *Mixtral:* Mixtral 8x22B. Best and second-best scores are **bolded** and <u>underlined</u>, respectively.

(a) Factuality Metrics

| Approach | Model | Essential-only | | | | | | Essential + Supplementary | | | | | |
| --- | --- | --- | --- | --- | --- | --- | --- | --- | --- | --- | --- | --- | --- |
| | | Macro | | | Micro | | | Macro | | | Micro | | |
| | | P | R | F1 | P | R | F1* | P | R | F1 | P | R | F1 |
| Together | Llama 3 | **74.2** | 38.6 | 47.3 | <u>65.3</u> | 32.8 | 43.7 | **89.2** | 32.4 | 42.2 | <u>83.7</u> | 22.9 | 35.9 |
| | Llama 4 | 67.0 | 47.5 | <u>53.1</u> | 62.4 | 41.8 | <u>50.0</u> | 81.1 | 39.0 | 48.8 | 79.1 | 28.8 | 42.2 |
| | Mixtral | 61.2 | 39.0 | 44.2 | 58.2 | 33.3 | 42.4 | 77.4 | 34.4 | 43.2 | 77.7 | 24.1 | 36.8 |
| Answer First | Llama 3 | <u>71.5</u> | 45.7 | 51.2 | 59.5 | 39.8 | 47.7 | 85.4 | 37.2 | 46.2 | 77.7 | 28.2 | 41.4 |
| | Llama 4 | 66.8 | <u>51.3</u> | **55.0** | 56.9 | **47.6** | **51.8** | 84.7 | <u>43.7</u> | **53.6** | 79.1 | <u>35.9</u> | **49.4** |
| | Mixtral | 60.3 | **51.6** | 52.6 | 49.7 | <u>45.9</u> | 47.7 | 77.9 | **45.6** | <u>53.4</u> | 71.7 | **36.0** | <u>47.9</u> |
| Evidence First | Llama 3 | 69.8 | 35.3 | 43.9 | **67.3** | 29.0 | 40.6 | <u>87.2</u> | 30.1 | 39.6 | **85.2** | 20.0 | 32.4 |
| | Llama 4 | 64.0 | 38.7 | 44.9 | 59.2 | 32.8 | 42.2 | 80.4 | 33.1 | 42.5 | 78.1 | 23.5 | 36.1 |
| | Mixtral | 55.1 | 38.5 | 41.7 | 61.5 | 32.1 | 42.2 | 71.2 | 34.7 | 41.4 | 80.2 | 22.7 | 35.4 |
| Heuristic | -- | 60.1 | 44.0 | 47.8 | 59.0 | 38.0 | 46.2 | 81.8 | 39.8 | 47.6 | 80.7 | 28.2 | 41.8 |

\* Essential-only Micro F1 is considered *Overall Factuality*.

(b) Relevance Metrics using Human Answers

| Approach | Model | BLEU | ROUGE | SARI | BERTScore | MEDCON | AlignScore | Overall Relevance |
| --- | --- | --- | --- | --- | --- | --- | --- | --- |
| Together | Llama 3 | 2.4 | 21.0 | 49.2 | 39.3 | 36.7 | **47.0** | 32.6 |
| | Llama 4 | 5.5 | 22.7 | 49.2 | 40.2 | 38.3 | 35.6 | 31.9 |
| | Mixtral | 5.7 | 23.1 | 51.8 | 40.2 | <u>39.9</u> | 38.3 | <u>33.2</u> |
| Answer First | Llama 3 | 3.4 | 22.7 | 51.0 | <u>40.4</u> | 38.9 | 41.5 | 33.0 |
| | Llama 4 | <u>6.7</u> | <u>23.6</u> | 51.7 | 40.3 | 38.1 | 36.0 | 32.7 |
| | Mixtral | **6.8** | **24.2** | <u>53.1</u> | **42.5** | **40.2** | 35.8 | **33.8** |
| Evidence First | Llama 3 | 1.7 | 20.0 | 49.1 | 37.4 | 31.3 | <u>43.6</u> | 30.5 |
| | Llama 4 | 4.0 | 20.3 | 48.8 | 36.9 | 32.8 | 37.4 | 30.0 |
| | Mixtral | 4.9 | 22.6 | 52.9 | 38.7 | 39.0 | 39.8 | 33.0 |
| Heuristic | -- | 6.7 | 22.1 | **53.6** | 30.4 | 35.0 | 34.0 | 30.3 |

(c) Relevance Metrics using Note Excerpt and Questions

| Approach | Model | BLEU | ROUGE | SARI | BERTScore | MEDCON | AlignScore | Overall Relevance |
|---|---|---|---|---|---|---|---|---|
| Together | Llama 3 | 0.3 | 17.8 | 49.4 | 23.7 | 29.0 | <u>59.0</u> | 29.9 |
| | Llama 4 | 1.8 | 25.1 | 49.0 | 28.5 | 36.9 | 54.7 | 32.7 |
| | Mixtral | 1.5 | 22.9 | 52.9 | 27.1 | 35.2 | 54.2 | 32.3 |
| Answer First | Llama 3 | 0.4 | 19.2 | 51.2 | 24.7 | 30.0 | 58.1 | 30.6 |
| | Llama 4 | <u>1.9</u> | <u>25.3</u> | 51.4 | 28.2 | 34.9 | 54.0 | 32.6 |
| | Mixtral | 1.6 | 24.9 | <u>54.2</u> | <u>28.9</u> | <u>37.9</u> | 52.9 | <u>33.4</u> |
| Evidence First | Llama 3 | 0.2 | 16.2 | 49.1 | 22.0 | 26.0 | 56.9 | 28.4 |
| | Llama 4 | 1.1 | 22.0 | 48.6 | 25.6 | 32.5 | 55.9 | 31.0 |
| | Mixtral | 1.4 | 22.7 | 53.0 | 26.5 | 34.5 | 54.3 | 32.1 |
| Heuristic | -- | **11.7** | **37.3** | **72.3** | **38.8** | **44.1** | **66.1** | **45.0** |

(d) Overall Scores. *Human:* computed with relevance metrics using Human Answers from (b); *Notes:* computed with relevance metrics using Note Excerpt and Questions from (c).

| Approach | Model | Overall Score (Human) | Overall Score (Notes) |
|---|---|---|---|
| Together | Llama 3 | 38.1 | 36.8 |
| | Llama 4 | <u>41.0</u> | 41.3 |
| | Mixtral | 37.8 | 37.3 |
| Answer First | Llama 3 | 40.3 | 39.2 |
| | Llama 4 | **42.3** | <u>42.2</u> |
| | Mixtral | 40.7 | 40.6 |
| Evidence First | Llama 3 | 35.5 | 34.5 |
| | Llama 4 | 36.1 | 36.6 |
| | Mixtral | 37.6 | 37.1 |
| Heuristic | -- | 38.2 | **45.6** |

# Discussion

We presented a benchmarking dataset, ArchEHR-QA, comprising patient questions, clinician-interpreted versions of questions, corresponding discharge summaries with sentence-level relevance annotations, and clinician-written answers. To our knowledge, there is no publicly available resource that coupled patient information needs from their EHRs with actual clinical documents. We achieve this by combining two different publicly accessible databases on a novel premise: "*what questions do patients ask about the clinical course of their hospitalization?*". With a total of 134 patient cases with detailed annotations, the proposed dataset will serve as a strong benchmark to develop and evaluate AI tools for EHR-specific information seeking.

During our manual error analysis of the highest and lowest scoring runs (based on Factuality and Relevance metrics) by the Llama 4 model using the Answer First approach, we observed that the automated metrics generally aligned with the actual quality of the generated responses. In most instances, the model-generated answers successfully captured the core content of the clinician-authored reference answers. For example, in one case, the model correctly inferred that the patient *"was given antibiotics because she developed pneumonia,"* but incorrectly attributed this decision to *"a chest x-ray that indicated low oxygen levels"*—a detail not supported by the clinical note and inconsistent with general medical knowledge. In another case, the model failed to mention an important related procedure (*"thoracentesis"*), despite its relevance to the question.

In one of the lowest-performing cases, the model incorrectly asserted that *"no specific order [was] mentioned for taking the prescribed medications,"* whereas the clinical note clearly included instructions for a *"dexamethasone taper."* Moreover, in the same instance, the model disproportionately reproduced much of the content from patient's question, rather than grounding the answer in clinical evidence. Interestingly, in some cases, although the model included citations to relevant sentences from the note, it failed to accurately convey the associated information in the generated text. Complete examples and corresponding annotations from our manual review are included in Supplementary Material (Table S2).

Our evaluation framework targeted two key dimensions: *Factuality*, assessing the cited note sentences, and *Relevance*, evaluating the answer text consistency with the reference response. Manual inspection showed alignment with these automated assessments. Nonetheless, we acknowledge the potential for more scalable and nuanced evaluations, such as leveraging advanced LLMs to assess answer quality, particularly in relation to annotated reference answers[44].

In addition to the rigorous validation protocol employed throughout the development of ArchEHR-QA, the dataset has also been subjected to community-based evaluation through a shared task (also titled ArchEHR-QA), organized as part of the BioNLP (Biomedical Natural Language Processing) Workshop at the 2025 Annual Meeting of the Association for Computational Linguistics (ACL 2025). This shared task invited researchers to explore and develop innovative methods for grounded EHR QA using the dataset. A total of 76 successful system submissions from 30 participating teams were recorded, further demonstrating the usability, relevance, and community interest in ArchEHR-QA as a benchmark resource.

The forum posts and MIMIC EHR notes originate from different patient populations. Nonetheless, the informational needs expressed in forum posts remain relevant and have been examined in contexts such as Radiology[45] and Laboratory Results[46,47]. Additionally, no forum post was aligned with a note in a way that could alter its underlying information needs. Patient portal messages linked to EHRs could offer real-world data to study patient information needs. However, although studies using portal messages exist, the corresponding datasets are not publicly available[16,17,48,49]. Our approximate but methodical alignment provides a publicly accessible resource for advancing patient-specific EHR QA.

The primary focus of studies on generating responses to patient portal questions has been on evaluating clinicians' perception rather than evaluating the use of EHR information[21,22]. The proposed dataset addresses this gap by providing detailed sentence-level EHR note annotations, enabling fine-grained assessments of how well automated systems ground their responses in relevant clinical information. Additionally, there is growing evidence supporting the use of AI technologies to generate draft responses for clinician review[21,22,50]. This dataset facilitates the development of automated benchmarking protocols, accelerating system improvements, and ultimately helping to reduce clinician burden.

The developed dataset originates from a single medical institution and a single health discussion forum. However, the underlying EHR data are drawn from two distinct clinical settings (ICU and ED) and were authored by different providers over several years. The associated patient questions also reflect authentic information needs expressed in real-world contexts. Moreover, our proposed paradigm for aligning forum posts with corresponding EHRs is scalable and adaptable to a variety of EHR systems, clinical documentation formats, and health discussion forums.

While the models are benchmarked, and demonstrate usability of the data, on the task of grounded EHR QA, the proposed dataset offers broader utility beyond this specific application. For instance, the annotated note excerpts can support the development and evaluation of clinical information retrieval systems[51]. Additionally, the clinician- authored questions represent a valuable resource for advancing research on summarizing patient questions[52]. Finally, the grounded QA task itself can also be extended to full-length clinical notes, whose document identifiers are preserved and will be released with the dataset, thus enabling investigations on long-form clinical QA[53].

# Conclusions

We introduced a novel dataset, ArchEHR-QA, with 134 real-world patient cases, each containing a patient-posed question, its clinician-interpreted equivalent, associated clinical note excerpts with sentence-level relevance annotations, and clinician-authored answers. We benchmarked this dataset using three state-of-the-art open-weight LLMs under a zero-shot setting across three distinct prompting strategies. Among these, the Answer First prompting approach consistently outperformed the others, and the Llama 4 model achieved the highest overall score of 42.3% (with a Factuality micro F1 score of 51.8% and an overall Relevance score of 32.7%). Given the substantial room for improvement identified by automated evaluation and manual error analysis, this dataset offers a valuable resource for advancing the development and evaluation of patient-focused EHR QA systems.

# Data and Code Availability

The proposed dataset, ArchEHR-QA, is publicly available through the controlled-access repository PhysioNet at https://doi.org/10.13026/zzax-sy62. The accompanying code used to evaluate the baseline systems is available via GitHub at https://github.com/soni-sarvesh/archehr-qa/tree/main/evaluation.

# Author Contributions

SS and DDF conceived the idea and developed the dataset. SS conducted the benchmarking experiments and wrote the initial draft of the manuscript. Both authors reviewed, edited, and approved the final version of the manuscript.

# Competing Interests

The authors declare no competing interests.

# Acknowledgements

This research was supported by the Division of Intramural Research (DIR) of the National Library of Medicine (NLM), National Institutes of Health, and utilized the computational resources of the NIH HPC Biowulf cluster (http://hpc.nih.gov). The content is solely the responsibility of the authors and does not necessarily represent the official views of the National Institutes of Health. The authors thank Dr. Saeed Arabi for his contributions to annotations and discussions of the evaluation parameters.

# Supplementary Materials

# Annotation Guidelines

## 1. Alignment

You will be provided with a patient-posed question from a public health discussion forum and a set of 10 candidate discharge summaries or notes, selected based on lexical and semantic similarity. Your task has two components:

(a) Note Selection for Alignment
- Select the discharge summary that is most appropriate for answering the question.
- The selected note should not contradict the question. While a perfect match is not expected, the alignment must be contextually plausible.
- Avoid selecting discharge summaries with mismatched core clinical facts.
    - Example: If the question is about the reason for a major procedure (e.g., CABG), and the note does not mention such a procedure, it is not a suitable match.

(b) Question Editing
- Edit the patient-posed question to reflect the patient characteristics and clinical context in the selected discharge summary.
- Make only the minimal changes required to ensure consistency between the question and the selected note without altering the core information need.
    - Example: If the question references a 56-year-old, but the note refers to a 58-year-old with otherwise matching context, change the age to 58 in the edited question.
- Do not add new clinical inquiries or modify the underlying intent of the question.
- Note: the clinical document is read only; it cannot be modified.

## 2. Clinician Question Creation and Sentence Relevance Annotations

Given an edited patient-posed question and its associated discharge summary, you will complete two sub-tasks:

(a) Clinician Question Formulation
- Identify the main information needs embedded in the patient's question.
- Reformulate it from the clinician's perspective to produce a clear and focused Clinician Question.
- This question should express how a clinician would interpret the patient's question and respond to it.
- Additionally, annotate the focus areas in the patient's question that prompted the Clinician Question.

(b) Sentence Relevance Annotation

- The discharge summary is pre-split into individual sentences.
- For each sentence, assign one of the following relevance labels indicating its importance in answering the Clinician Question:
    - Essential: Provides critical and necessary information to directly answer the Clinician Question. Note: if several sentences contain the same critical information, all should be labeled as essential.
    - Supplementary: Offers supporting or contextual information but is not strictly required.
    - Not-Relevant: Does not contribute meaningfully to answering the question. This is the default label unless updated.

## 3. Answer Creation

Given the patient-posed question with focus areas, the clinician-interpreted question, and the annotated discharge summary with sentence-level relevance annotations, you will compose an answer. The following are the guidelines to create an answer:

- Write a succinct answer (approximately 75 words or 5 sentences) that directly addresses the Clinician Question.
- The answer must be fact-based and entirely grounded in the discharge summary.
    - Do not introduce or infer information that is not explicitly stated in the discharge summary.
- All Essential sentences must be cited in your answer using bracketed numeric references (e.g., "[1]", "[2,3]").
    - Supplementary sentences may be cited for clarity or completeness but are not required.
- Ensure that the answer:
    - Fully addresses all aspects of the Clinician Question.
    - Is concise, medically accurate, and free of speculation.
    - Is written in a neutral, informative tone, suitable for a clinical audience (layperson language is not required).

# Results Analysis

**Table S1:** System-generated responses to the example case presented in the paper.

| Approach | Model | Answer |
|---|---|---|
| Together | Llama 3 | The emergency salvage repair was performed due to a ruptured thoracoabdominal aortic aneurysm. |1,2| The surgery involved a 34-mm Dacron tube graft and deep hypothermic circulatory arrest. |2| |
| Together | Llama 4 | They performed the emergency salvage repair because the patient had a ruptured thoracoabdominal aortic aneurysm. |1,2| The surgery was necessary to address the rupture, which required immediate attention, and involved a 34-mm Dacron tube graft using deep hypothermic circulatory arrest. |2| The procedure was done to save the patient's life after he experienced cardiac arrest twice during the operation. |3| |
| Together | Mixtral | The emergency salvage repair was performed due to a ruptured thoracoabdominal aortic aneurysm. |1| This was an urgent procedure involving a 34-mm Dacron tube graft and deep hypothermic circulatory arrest. |2| |
| Answer First | Llama 3 | The patient underwent emergency salvage repair due to a ruptured thoracoabdominal aortic aneurysm. |1,2| This surgery was necessary to prevent further bleeding and potential death. |
| Answer First | Llama 4 | The emergency salvage repair was performed because the patient had a ruptured thoracoabdominal aortic aneurysm. |1,2| The surgery was necessary to address the rupture, which likely posed an immediate threat to the patient's life. |1| The procedure involved a 34-mm Dacron tube graft and deep hypothermic circulatory arrest due to the complexity of the case. |2| |
| Answer First | Mixtral | The surgery was performed due to a ruptured thoracoabdominal aortic aneurysm. |1| This condition is life-threatening and requires immediate intervention. The procedure involved an emergent salvage repair using a 34-mm Dacron tube graft to prevent further rupture and restore normal blood flow. |2| |
| Evidence First | Llama 3 | The emergency salvage repair was performed due to a ruptured thoracoabdominal aortic aneurysm. |1,2| The surgery involved a 34-mm Dacron tube graft and deep hypothermic circulatory arrest. |2| |
| Evidence First | Llama 4 | They performed the emergency salvage repair because the patient had a ruptured thoracoabdominal aortic aneurysm. |1,2| The surgery was necessary to address the rupture, which required immediate attention, and involved a 34-mm Dacron tube graft using deep hypothermic circulatory arrest. |2| The procedure was done to save his life after he experienced cardiac arrest twice during the operation. |3| |

| | Mixtral | Your father underwent an emergency salvage repair. |1| This was due to a ruptured thoracoabdominal aortic aneurysm. |1| The procedure involved using a 34-mm Dacron tube graft with deep hypothermic circulatory arrest. |2| |
|---|---|---|
| Heuristic | -- | He was transferred to the hospital on 2025-1-20 for emergent repair of his ruptured thoracoabdominal aortic aneurysm. |1| He was taken immediately to the operating room where he underwent an emergent salvage repair of ruptured thoracoabdominal aortic aneurysm with a 34-mm Dacron tube graft using deep hypothermic circulatory arrest. |2| Postoperatively he was taken to the intensive care unit for monitoring with an open chest. |4| |

**Table S2:** System-generated responses by Llama 4 using the Answer First approach on the two best and worst performing cases on Overall Score. Note Excerpts are not included because of brevity and MIMIC data use agreement.

(a) Example

| Case ID: 23 (best factuality) | |
|---|---|
| **Patient Question** | Hi Dr. My mom is 88 years old, no heart problems. She hit her head and was hospitalized. No intubation was done in ICU. Her blood count and pressure were normal, she could eat and was breathing on her own. <u>My question is about the antibiotics she was given during her hospital stay. Were they necessary?</u> Thank you, |
| **Clinician Question** | Why was she given antibiotics during her hospital stay? |
| **Clinician Answer** | The patient was given antibiotics during her hospital stay because of a lung infection [5]. The patient was noted to have low levels of oxygen and a chest x-ray was ordered [5]. It indicated the patient was developing pneumonia [5]. IV antibiotics were initiated to treat the pneumonia [6]. The patient showed improvement right away after starting the antibiotics [6]. Therefore, the antibiotics were necessary during her hospital stay. |
| **System Answer** | ```answer. Your mom was given antibiotics because she developed pneumonia, as shown by a chest x-ray that indicated low oxygen levels. |5,6| The antibiotics were started immediately and she showed improvement. |6| ``` |
| **Manual Review** | • Answers the question?<br>  ○ Yes: The antibiotics were initiated because of pneumonia.<br>• Uses the evidence?<br>  ○ Supported: The antibiotics were initiated because of pneumonia.<br>  ○ Refuted: The chest x-ray was ordered due to low oxygen levels, not the other way around.<br>• Uses general knowledge?<br>  ○ Conflicting: A chest x-ray does not measure oxygen saturation.<br>• Formatting<br>  ○ Uses additional "answer" markdown block in the response. |
| Case ID: 39 (best relevance) | |
| **Patient Question** | hello, all doctors i am here for my mom is suffering from liver cirrhosis and stone in gall bladder, a few days back she felt unconcious then we took her to hospitol then she was kept in ICU for 10 days then after treatment she came back to censes, in the mean process aprox 1 and half litre water pumped from chast and now again <u>there is sign of swelling and filling of water in there</u> then again we consult our doctor they suggested for liver transplant,here now I am asking from you people that <u>what is the best and proper way to cure her=.</u> |

| Clinician Question | Is she getting any treatment for the swelling? |
|---|---|
| Clinician Answer | The patient was found to have a pleural effusion caused by liver disease [7]. The treatment she received for the swelling consisted of thoracentesis to fix the fluid accumulation [8]. The patient will continue on diuretics, lasix 10 mg daily and spirolactone 50 mg daily to help prevent a reoccurrence of the pleural effusion [9,14]. The patient will continue to follow up with Hepatology as indicated [16]. Her urine output and weight will be monitored and adjusted as needed [15]. |
| System Answer | She is currently on diuretics, specifically lasix and spironolactone, to help prevent a reoccurrence of the pleural effusion. \|9\| Her diuretics were briefly held but have been restarted at lasix 10mg daily and spironolactone 50mg daily. \|13,14\| Her urine output and weight will continue to be monitored and adjusted as needed. \|15\| |
| Manual Review | <ul><li>Answers the question?<ul><li>Partially: It accurately describes the current medication regimen (diuretics), mentions the specific drugs and dosages, and notes the monitoring of urine output and weight. However, it does not mention thoracentesis and the plan to follow up with Hepatology.</li></ul></li><li>Uses the evidence?<ul><li>Supported: Uses appropriate evidence for the included clinical information.</li></ul></li><li>Uses general knowledge?<ul><li>No: The system does not incorporate additional general knowledge beyond what is included in the input note.</li></ul></li></ul> |
| **Case ID: 118 (least relevance)** ||
| Patient Question | patient has been operated because of intestine, patient had scepticimia and surgery again for cleaning inside, what are the chances of survival and what can be the consequences after all these operations ?? |
| Clinician Question | What is the expected course of recovery after surgery and septicemia? |
| Clinician Answer | The patient is expected to continue on antibiotics via her PICC line when she is discharged, due to GNRs and GPCs that were present in the blood cultures [18,19,23,36]. Surgery included hysterectomy, salpingo-oopherectomy and omentectomy, resulting in a colostomy [9,23,33]. The patient will have an extended recovery as her abdominal wounds heal, and as she learns how to manage her colostomy [26,34,35]. The patient will be discharged to a rehab facility for extended recovery [24]. |
| System Answer | ```answer. The patient's condition improved at discharge, with stable vital signs and decreasing white blood cell count. \|26,35\| She tolerated a regular diet, ambulated with assistance, and had well-controlled pain. \|26\| Expected course of recovery involves continued improvement with rehabilitation. \|24,25\| Close monitoring and management of potential complications, such as infection, will be necessary. \|18,21,23\| Underlying conditions like DM2, HTN, and HL require ongoing management. \|2\| ```. |
| Manual Review | <ul><li>Answers the question?<ul><li>Partially: The system output presents a general summary of the patient's discharge condition and some aspects of the expected recovery. However, it fails to mention the continuation of antibiotics due to GNRs/GPCs, colostomy care, and wound healing.</li></ul></li><li>Uses the evidence?<ul><li>No: The citations used for infection management (\|18,21,23\|) are not explicitly tied to the continuation of antibiotics.</li></ul></li><li>Uses general knowledge?<ul><li>Yes: Uses general medical knowledge in mentioning the need for monitoring complications like infection and managing chronic conditions such as DM2, HTN, and HL.</li></ul></li></ul> |
| | • Formatting<br>    o Uses additional "answer" markdown block in the response. |
| **Case ID: 99 (least factuality)** ||
| **Patient Question** | Hi Doc my husband has been identified with tuburculos meningitis as his protien was 496 / lympocytes 90% after lumber tap..he spent 5 days in ICU & now he is on akt 4 treatment. 01 Large Tablet after breakfast / rifamicin after 02 hours of breakfast and rest two after lunch is this the right way to take the medicines as earlier he was taking rifamicin empty stomach but this creates nausea and vomiting so our doc changed to the sequel mentioned. Also he has a fever around 100 - 101 please advise |
| **Clinician Question** | Is there a specified order to take the prescribed medications? |
| **Clinician Answer** | Patient was diagnosed with tuberculous meningitis and was discharged home on a long list of important medications that must be taken exactly as directed [2,3]. The dexamethasone taper medication has a specific order in which it is to be taken [21]. Patient is to start at 6mg IV every 6 hours and decrease the dose over a period of time until transitioned to an oral dose [21]. This dose starts at 4mg daily and gradually decreases until finished [21]. |
| **System Answer** | There is no specific order mentioned for taking the prescribed medications. \|4,6,9,10,11\| However, the patient is currently taking one large tablet (likely Isoniazid) after breakfast, Rifampin after 2 hours of breakfast, and the rest two tablets after lunch. The current regimen seems to be adjusted to mitigate nausea and vomiting caused by Rifampin taken on an empty stomach. It is recommended to consult the doctor for further clarification on the medication schedule. \|24,27,28\| |
| **Manual Review** | • Answers the question?<br>    o No: It states that there is no specific order mentioned, which is incorrect. It fails to address the question regarding the specific order of taking the prescribed medications, particularly the dexamethasone taper.<br>• Uses the evidence?<br>    o Refutes: It misrepresents the evidence by claiming that no specific order is mentioned. Further, it focuses on observed patient behavior (timing of pill intake from the question) rather than the prescribed regimen described in the evidence.<br>• Uses general knowledge?<br>    o No: The system does not incorporate additional general knowledge beyond what is included in the input question and the note. |

(b) Factuality Metrics

| Case ID | Strict | | | Lenient | | |
|---|---|---|---|---|---|---|
| | P | R | F1 | P | R | F1 |
| 23 (best factuality) | 100 | 100 | 100 | 100 | 100 | 100 |
| 39 (best relevance) | 75.0 | 50.0 | 60.0 | 75.0 | 50.0 | 60.0 |
| 118 (least relevance) | 62.5 | 50.0 | 55.6 | 75.0 | 35.3 | 48.0 |
| 99 (least factuality) | 0 | 0 | 0 | 62.5 | 31.3 | 41.7 |

(c) Relevance Metrics using Human Answers

| Case ID | BLEU | ROUGE | SARI | BERTScore | MEDCON | AlignScore | Overall Relevance |
|---|---|---|---|---|---|---|---|
| 23 (best factuality) | 0 | 29.2 | 43.5 | 43.7 | 90.9 | 56.0 | 43.9 |
| 39 (best relevance) | 26.2 | 41.9 | 58.7 | 61.2 | 85.7 | 66.5 | 56.7 |
| 118 (least relevance) | 0 | 14.8 | 38.5 | 24.8 | 0 | 19.0 | 16.2 |
| 99 (least factuality) | 0 | 12.3 | 46.7 | 21.6 | 25.0 | 1.5 | 17.9 |

(d) Relevance Metrics using Note Excerpts and Questions

| Case ID | BLEU | ROUGE | SARI | BERTScore | MEDCON | AlignScore | Overall Relevance |
|---|---|---|---|---|---|---|---|
| 23 (best factuality) | 0 | 23.5 | 56.6 | 29.2 | 52.6 | 66.4 | 38.1 |
| 39 (best relevance) | 3.1 | 31.9 | 77.1 | 43.1 | 46.2 | 72.4 | 45.6 |
| 118 (least relevance) | 1.6 | 24.8 | 40.2 | 22.7 | 18.2 | 44.3 | 25.3 |
| 99 (least factuality) | 0 | 18.4 | 44.5 | 9.3 | 38.5 | 56.3 | 27.8 |

(e) Overall Scores

| Case ID | Overall Score (Human) | Overall Score |
|---|---|---|
| 23 (best factuality) | 71.9 | 69.0 |
| 39 (best relevance) | 58.4 | 52.8 |
| 118 (least relevance) | 35.9 | 40.4 |
| 99 (least factuality) | 8.9 | 13.9 |

# Other Details

## Keywords used to filter the health forum discussion posts

"admit", "admitted", "admission", "admitting", "admits", "icu", "ccu", "intensive care", "critical care", "care unit", "er", "emergency room", "ed", "emergency department", "emergency dept", "emergency"

**Table S3:** Prompt used to generate ICD code descriptions for forum posts.

---

Generate the relevant ICD codes for the health forum message [MESSAGE] from a patient:

[MESSAGE]
{message}

The output should be a markdown code snippet formatted in the following schema, including the leading and trailing "```icd" and "```":

```icd
[ICD Code] - [Description Text]
[ICD Code] - [Description Text]
[ICD Code] - [Description Text]
```